\documentclass[runningheads, envcountsame, a4paper]{llncs}

\usepackage{graphicx}
\usepackage{amsmath}
\usepackage{amssymb}
\usepackage{booktabs}
\usepackage{multirow}
\usepackage{array}
\usepackage{placeins}  
\usepackage{float}     
\usepackage{microtype} 
\usepackage{url}
\usepackage{hyperref}

\begin{document}


\title{MADP: A Multi-Agent Pipeline for Sustainable Document Processing with Human-in-the-Loop}

\titlerunning{MADP: Multi-Agent Document Processing}

\author{Diego Gosmar\inst{1}\orcidID{0009-0008-7513-1255} \and Giovanni Zenezini\inst{2}\orcidID{0000-0002-0996-6739}}

\authorrunning{D. Gosmar and G. Zenezini}

\institute{
	Tesisquare Head of AI, Turin, Italy; Member, Open Voice Interoperability Initiative, Linux Foundation AI \& Data\\
	\email{diego.gosmar@ieee.org}
	\and
	Polytechnic University of Turin, Department of Management and Production Engineering, Torino, Italy\\
	\email{giovanni.zenezini@polito.it}
}

\maketitle

\begin{abstract}
Document processing automation remains a critical challenge in enterprise environments, where traditional manual approaches are labor-intensive and error-prone. We present MADP, a multi-agent architecture that addresses the challenge of automating document processing in enterprise settings by combining deep learning-based classification and parsing with large language model extraction, while maintaining accuracy through selective human validation. Our system integrates five specialized agents—Classificator, Splitter, Parser, Extraction, and Validator—with a Human-in-the-Loop (HITL) mechanism and a novel Prompt Fine Tuning with Feedback Inheritance (PFTFI) approach. The operational analysis on a production use-case scenario of 100,000 invoices per year indicates a potential reduction of Full-Time Equivalent (FTE) requirements by approximately 70\%. Production deployment on 955 real-world documents processed through January 2026 achieves a 97.0\% full-pipeline automation rate, with only 3\% requiring non-AI fallback. Ablation evaluation on a stratified 100-document subset (5 documents per each of 20 supplier/document-type categories) demonstrates that the full MADP configuration with Human-in-the-Loop supervision attains 98.5\% document-level accuracy. Additionally, we present a comprehensive sustainability analysis showing that our hybrid AI+HITL approach reduces CO\textsubscript{2} emissions by 69\%, energy consumption by 69\%, and water usage by 63\% compared to traditional manual processing. Benchmark comparisons of multiple LLM backends (Granite-Docling, Mistral-Small, DeepSeek-OCR) provide practical insights for deployment in production environments.

\end{abstract}

\section{Introduction}

The digital transformation of document-intensive business processes presents significant challenges in information extraction, validation, and automated decision-making~\cite{wang2024docparsing}. Traditional Optical Character Recognition (OCR) systems struggle with complex document layouts, multi-page structures, and domain-specific terminology, often requiring extensive manual verification~\cite{belzarena2025ocr}.

Recent advances in Large Language Models (LLMs) have demonstrated remarkable capabilities in understanding and extracting structured information from unstructured documents~\cite{brown2020llm}. However, deploying LLMs in production environments faces three critical challenges: (1) hallucinations and non-deterministic accuracy concerns in mission-critical applications~\cite{zhang2024hallucination,gosmar2025hallucination,gosmar2026icaart}, (2) computational and environmental costs of large-scale inference~\cite{patterson2024carbon}, and (3) lack of interpretability and auditability in regulated industries~\cite{raji2024aiaudit}.

To address these challenges, we propose MADP, a multi-agent architecture that orchestrates specialised AI agents under human oversight for end-to-end document processing. The framework combines convolutional neural networks and large language models within a modular pipeline and introduces a Prompt Fine Tuning with Feedback Inheritance (PFTFI) mechanism that leverages human corrections to refine extraction behaviour over time without retraining the underlying models. We evaluate MADP on a production dataset of 955 real-world documents processed through January 2026, achieving a 97.0\% full-pipeline automation rate. Ablation evaluation on a stratified 100-document subset (5 documents per each of 20 supplier/document-type categories) demonstrates that the full configuration with Human-in-the-Loop supervision attains 98.5\% document-level accuracy. We further analyze operational efficiency through a use-case scenario involving 100{,}000 invoices per year, demonstrating a potential FTE reduction of approximately 70\%. In addition to operational metrics, we provide what is, to the best of our knowledge, the first detailed sustainability analysis of AI-assisted document processing, quantifying reductions in CO\textsubscript{2} emissions, energy consumption, and water usage relative to a fully manual baseline. Finally, we report a comparative benchmark of multiple LLM backends for document analysis tasks, highlighting the trade-offs between accuracy, latency, and resource footprint in production-like conditions.

Our results demonstrate that intelligent orchestration of AI agents with strategic human intervention achieves superior accuracy while significantly reducing both operational costs and environmental impact compared to fully manual or fully automated approaches.

\section{Related Work}

\subsection{Document Analysis and Recognition}

Traditional document analysis relies on rule-based systems and classical machine learning approaches~\cite{nagy2000twentyfive}. Deep learning has transformed the field, with CNNs achieving state-of-the-art performance in document classification~\cite{harley2015icdar} and layout analysis~\cite{xu2020layoutlm}. LayoutLM~\cite{xu2020layoutlm} and its successors combine text, layout, and visual features for document understanding, while Donut~\cite{kim2022donut} proposes an OCR-free approach using vision transformers.

Recent work on document parsing has focused on handling complex layouts~\cite{zhao2024doclayoutyolo}. ParseBench~\cite{zhang2026parsebench} evaluates 14 parsing methods on ${\sim}$2,000 enterprise pages across five capability dimensions, finding that no single parser dominates across document types — a result consistent with our pragmatic choice of Docling. For invoice-specific extraction, Liu et al.~\cite{liu2024matrix} propose MATRIX, a memory-augmented agent that outperforms direct LLM prompting by over 30\% on 764 real invoice documents; unlike MADP, MATRIX does not incorporate HITL validation or sustainability analysis. Most approaches lack robustness to diverse document formats and require extensive training data for domain adaptation.

\subsection{Multi-Agent Systems for NLP}

Multi-agent systems decompose complex tasks into specialized sub-tasks handled by coordinated agents~\cite{wu2023autogen,hong2023metagpt,gosmar2025hallucination,gosmar2026icaart}. AutoGEN~\cite{wu2023autogen} enables building LLM applications through conversational agents, while MetaGPT~\cite{hong2023metagpt} incorporates human workflows into agent orchestration. Emerging open standards such as the Open Voice Interoperability specifications~\cite{ovonspec} further address agent communication interoperability across heterogeneous systems. Kulkarni and Kulkarni~\cite{kulkarni2026multiagent} benchmark four orchestration architectures on 10,000 SEC filings, finding that reflexive self-correcting loops achieve the highest F1 (0.943) but at 2.3$\times$ the cost of sequential baselines; MADP's PFTFI achieves analogous iterative correction while controlling costs through selective HITL intervention.

In document processing, recent benchmarks for OCR and document understanding~\cite{liu2023ocrbench} have revealed performance gaps in handling multilingual and handwritten text, while systematic evaluation on production-scale datasets and environmental impact assessment remains limited.

\subsection{Human-in-the-Loop Systems}

Human-in-the-Loop (HITL) approaches combine automated processing with strategic human intervention~\cite{mosqueira2023hitl}. Active learning frameworks~\cite{settles2012active} minimize labeling costs by selectively querying human annotators for uncertain predictions. However, existing HITL systems for document processing lack mechanisms for continuous learning from human feedback.

\subsection{Sustainability of AI Systems}

The environmental impact of AI has received increasing attention~\cite{patterson2024carbon,strubell2019energy}. Recent work quantifies the carbon footprint of LLM training~\cite{luccioni2023power} and inference~\cite{samsi2023carbon}, as well as water consumption for data center cooling~\cite{li2024water}. However, no prior work has analyzed the full sustainability impact of AI-assisted document processing compared to manual alternatives.

\section{MADP Architecture}

\subsection{System Overview}

\begin{figure}[h!]
	\centering
	\vspace{-2mm}
	\includegraphics[width=0.55\textwidth]{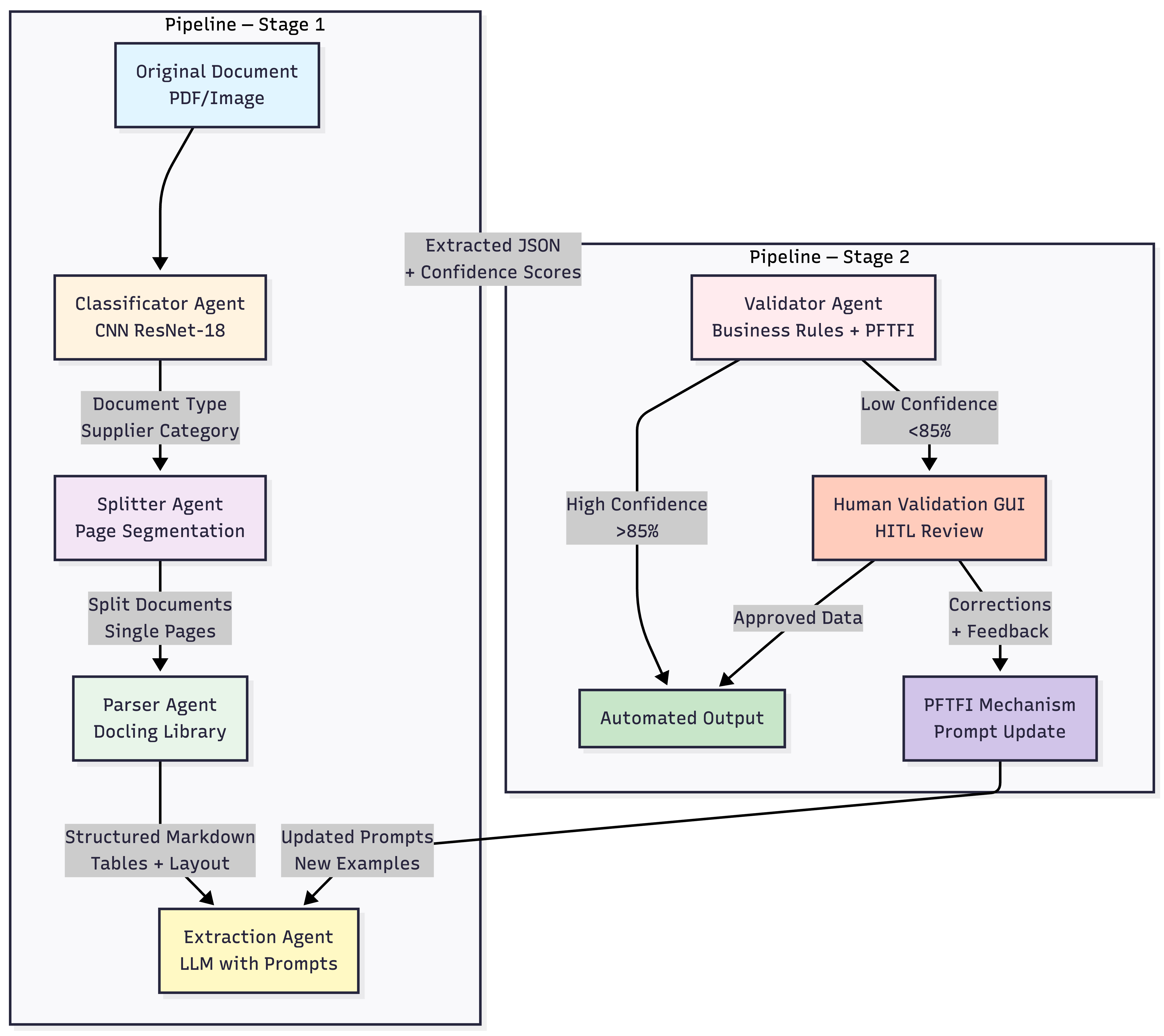}
	\caption{MADP pipeline: five sequential components (Classificator, Splitter, Parser, Extraction, Validator) with PFTFI feedback loop.}
	\label{fig:architecture}
	\vspace{-2mm}
\end{figure}

MADP implements a specialized-module pipeline with five components orchestrated sequentially (Figure~\ref{fig:architecture}). Each component performs a specific subtask, with intermediate results passed to downstream stages. The PFTFI feedback loop introduces bidirectional agentic behaviour across the pipeline, enabling continuous improvement without model retraining. The architecture provides inherent security benefits through validation at each stage~\cite{gosmar2025promptinjection} and supports both fully automated processing and Human-in-the-Loop validation at critical decision points.

\subsection{Classificator Agent}

The Classificator Agent uses a Convolutional Neural Network trained to identify document types and supplier categories. For invoice processing, the CNN classifies documents by supplier, enabling supplier-specific extraction templates.

The CNN architecture uses ResNet-18~\cite{he2016resnet} pretrained on ImageNet. ResNet-18 was selected deliberately over deeper variants (ResNet-50, ResNet-101) to minimise classification latency and computational footprint, while maintaining sufficient representational capacity for document header classification. The first three convolutional blocks remain frozen to preserve general visual features, while the fourth block is fine-tuned on document headers. For invoice and delivery note processing, input images are cropped to the top 40\% to focus on header regions containing supplier identification marks, logos, and document type indicators. Images are resized to 224×224 pixels with standard normalization. The model achieves 95.3\% classification accuracy on our test set of 5,000 documents across 150 supplier categories.

\subsection{Splitter Agent}

Multi-page documents require page-level segmentation before processing. The Splitter Agent analyzes document structure and separates logical units (e.g., individual invoices within a batch file). It uses a combination of visual features (page breaks, headers) and semantic analysis to identify boundaries.

For PDF documents, the agent extracts page metadata and applies heuristics based on document type classification. The splitter reduces processing errors caused by context mixing across document boundaries.

\subsection{Parser Agent}

The Parser Agent converts raw document formats into structured markdown representations that are optimised for subsequent LLM-based processing. This preprocessing step is critical for the observed accuracy improvements, as it reduces noise and exposes the underlying document structure in a form that is easier for the extraction models to exploit. Concretely, the parser analyses the page layout to identify text regions, tables, and figures, determines an appropriate reading order to reconstruct a coherent textual flow, and recognises table structures at the level of individual cells. The resulting content is then transformed into a hierarchical markdown representation that preserves key spatial relationships while substantially reducing token count relative to raw OCR output.

We implemented the parser using the Docling library~\cite{docling2024}, which combines rule-based layout analysis with ML-based table detection. The structured markdown format reduces token count by 35\% compared to raw OCR output while preserving semantic information.

\subsection{Extraction Agent}

The Extraction Agent employs large language models to infer structured information from the parsed document representations. Extraction is driven by carefully designed prompts that specify the document type, the set of expected fields, and the desired output format, together with explicit constraints on data types and admissible values. The prompt also incorporates a small number of illustrative examples that demonstrate correct extractions for representative cases, and it includes guidance on how to handle missing, ambiguous, or partially inconsistent information so as to minimise hallucinations and enforce schema compliance.

The agent outputs JSON-structured data with confidence scores for each extracted field. We evaluate multiple LLM backends (Section~\ref{sec:llm-benchmark}) to balance accuracy, latency, and cost.

\subsubsection{Parallel Extraction Strategy}

To increase robustness and confidence scores, the Extraction Agent supports parallel extraction configurations where multiple LLM backends or specialised models process identical documents concurrently. The system applies consensus voting to identify agreed-upon values, which receive elevated confidence scores, while discrepancies trigger automatic flagging for human review. Field-specific extractors may be deployed in parallel for complex structures such as line-item tables, handwritten annotations, or multilingual content, enabling targeted optimization without compromising overall pipeline throughput.

\subsection{Validator Agent with PFTFI}

The Validator Agent performs post-processing validation and implements the proposed Prompt Fine Tuning with Feedback Inheritance (PFTFI) mechanism. This component acts as a sentinel~\cite{gosmar2025sentinel}, enforcing data quality and security through systematic verification of the extracted fields against domain knowledge and business constraints. In practice, the validator checks that dates, monetary amounts, and tax identifiers conform to the expected formats; it verifies that numerical relationships such as the equality between subtotal plus tax and total amount hold within predefined tolerances; it enforces plausible value ranges for key quantities; and it inspects cross-field dependencies to detect internal inconsistencies before the data are propagated to downstream systems.

Documents failing validation or with confidence scores below configurable target thresholds are routed to human reviewers through the validation GUI. These thresholds are typically set between 80-90\% based on document type, supplier reliability, field criticality, and organizational risk tolerance, and can be adjusted dynamically based on observed accuracy patterns.

When humans correct extraction errors, the PFTFI agent captures the original extracted data alongside the corrections, generates structured feedback that describes the error pattern, updates both the extraction and parsing prompts by incorporating new examples and refining instructions, and applies these corrections to pending documents through a feedback inheritance mechanism.

\begin{figure}[h!]
	\centering
	\vspace{-2mm}
	\includegraphics[width=0.72\textwidth]{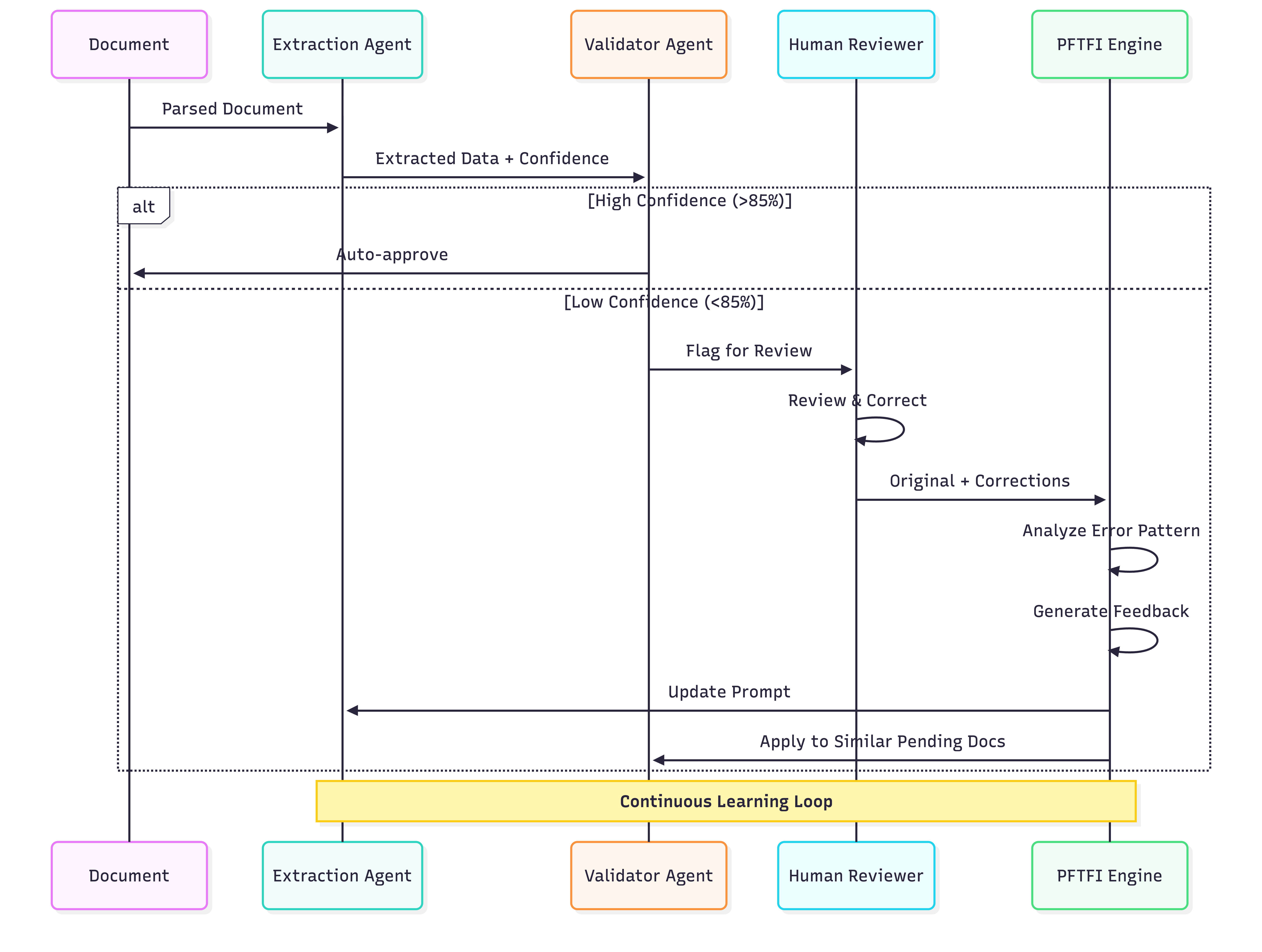}
	\caption{PFTFI mechanism: human corrections update extraction prompts and propagate to pending documents without model retraining.}
	\label{fig:pftfi}
	\vspace{-2mm}
\end{figure}

The complete PFTFI workflow is shown in Figure~\ref{fig:pftfi}.

Updated prompts are versioned and propagated bidirectionally through the pipeline. Semantic corrections refine Extraction Agent prompts by incorporating new examples and constraints, while layout-related feedback such as table structure misinterpretation or reading order errors updates Parser Agent configuration. This dual feedback loop enables continuous improvement across both structural parsing and semantic extraction stages without requiring model retraining. The PFTFI mechanism applies corrections not only to future documents but also to pending documents in the processing queue that exhibit similar characteristics through a feedback inheritance mechanism.

\subsection{Human Validation Interface}
The validation interface presents each document side by side with the corresponding extracted metadata in structured JSON format, enabling efficient human inspection and correction. The design highlights fields with low confidence scores, provides compact interaction mechanisms for amendments such as single-click corrections and constrained selection menus, and visualises differences between the original document content and the inferred metadata values. For documents that meet predefined confidence and consistency thresholds, the interface supports rapid confirmation without granular inspection.

Human operators correct the extracted metadata fields rather than annotating source documents directly. Corrections are captured as structured feedback that includes the original value, corrected value, field name, document type, and supplier category, providing rich signal for the PFTFI mechanism. In the current deployment, the average human review time is approximately 45 seconds per flagged document, compared to around 120 seconds for fully manual processing of the same invoices.

\subsubsection{Atomic Consistency Checks}

Beyond format validation, the Validator Agent implements atomic consistency checks that leverage internal document relationships to verify data correctness without external references. For invoice processing, these deterministic validations include arithmetic constraints where subtotal plus tax equals total amount within predefined tolerances, quantity reconciliation ensuring line item sums match header totals, cross-field consistency such as invoice date preceding due date, and domain constraints verifying that VAT rates conform to legal values and currency codes match ISO 4217 standards.

When atomic checks pass, the system automatically elevates confidence scores to near certainty for the validated fields, substantially reducing human review burden even when initial LLM extraction confidence was moderate. This approach effectively combines symbolic reasoning with neural extraction, creating a hybrid validation layer that catches errors deterministically before they propagate to downstream systems.

\section{Experimental Evaluation}

\subsection{Dataset and Evaluation Metrics}

We evaluate MADP accuracy and operational metrics on a production dataset comprising 955 documents processed through January 2026. For FTE reduction analysis and sustainability impact assessment, we use a representative use-case scenario of 100{,}000 invoices per year based on typical enterprise workloads. The 955-document evaluation set covers the same diversity characteristics as the larger use-case scenario: more than 50 distinct suppliers with heterogeneous invoice templates, documents in Italian, English, German, and French, and complex layouts ranging from single-page invoices to multi-page statements, reflecting the diversity encountered in real operational settings. Experiments were also conducted using Arabic-language documents for two sample supplier use cases.

The evaluation focuses on several complementary metrics. Document-level accuracy measures the proportion of invoices for which all required fields are extracted correctly. Field-level F\textsubscript{1} scores capture the balance between precision and recall at the level of individual data items. We also report end-to-end processing time per document, which includes all stages of the pipeline, and the human intervention rate, defined as the fraction of documents that require manual review through the validation interface. Ground truth labels are obtained via dual human annotation, with disagreements resolved through an adjudication process to ensure high-quality reference data.

\subsection{Production Deployment Results}
\label{sec:production}

We report operational statistics from production deployment, extending through January 2026 (Table~\ref{tab:production-stats}). During this period, the system processed a total of 955 documents, of which 926 (97.0\%) were handled end-to-end by the MADP pipeline without requiring fallback to alternative workflows. The remaining 29 documents (3.0\%), comprising 9 routed to legacy OCR systems and 20 processed via a manual web portal, represent cases in which unrecognised formats, severe image degradation, or explicit operator override triggered non-AI processing paths. These 29 documents are excluded from the accuracy evaluation as they were deliberately routed outside the MADP pipeline and do not represent processing errors.
\begin{table}[!htbp]
	\centering
	\caption{Production deployment statistics (data up to 21 January 2026, 16:00): documents processed by MADP versus non-AI fallback workflows}
	\label{tab:production-stats}
	\begin{tabular}{lcc}
		\toprule
		\textbf{Processing Method} & \textbf{Count} & \textbf{Percentage} \\
		\midrule
		AI-based (MADP pipeline) & 926 & 97.0\% \\
		Non-AI (OCR + Manual portal) & 29 & 3.0\% \\
		\midrule
		Total documents processed & 955 & 100\% \\
		\bottomrule
	\end{tabular}
\end{table}
The observed 97\% AI-based automation rate compares favourably to reported benchmarks for traditional OCR-based invoice processing systems, which typically achieve 85–95\% accuracy on structured documents~\cite{parseur2025benchmark} and often require manual intervention on over 20\% of cases~\cite{turian2025ai}. This confirms that the multi-agent architecture generalises effectively to documents encountered in daily operations and that the HITL validation interface allows selective human intervention on edge cases while preserving throughput for the majority of invoices. The deployment experience also underscores the value of modular design: when a document fails at the classification or parsing stage, the system degrades gracefully to manual inspection rather than producing unreliable extractions.

\subsection{Ablation Study}

\begin{figure}[h!]
\centering
\vspace{-2mm}
\includegraphics[width=0.68\textwidth]{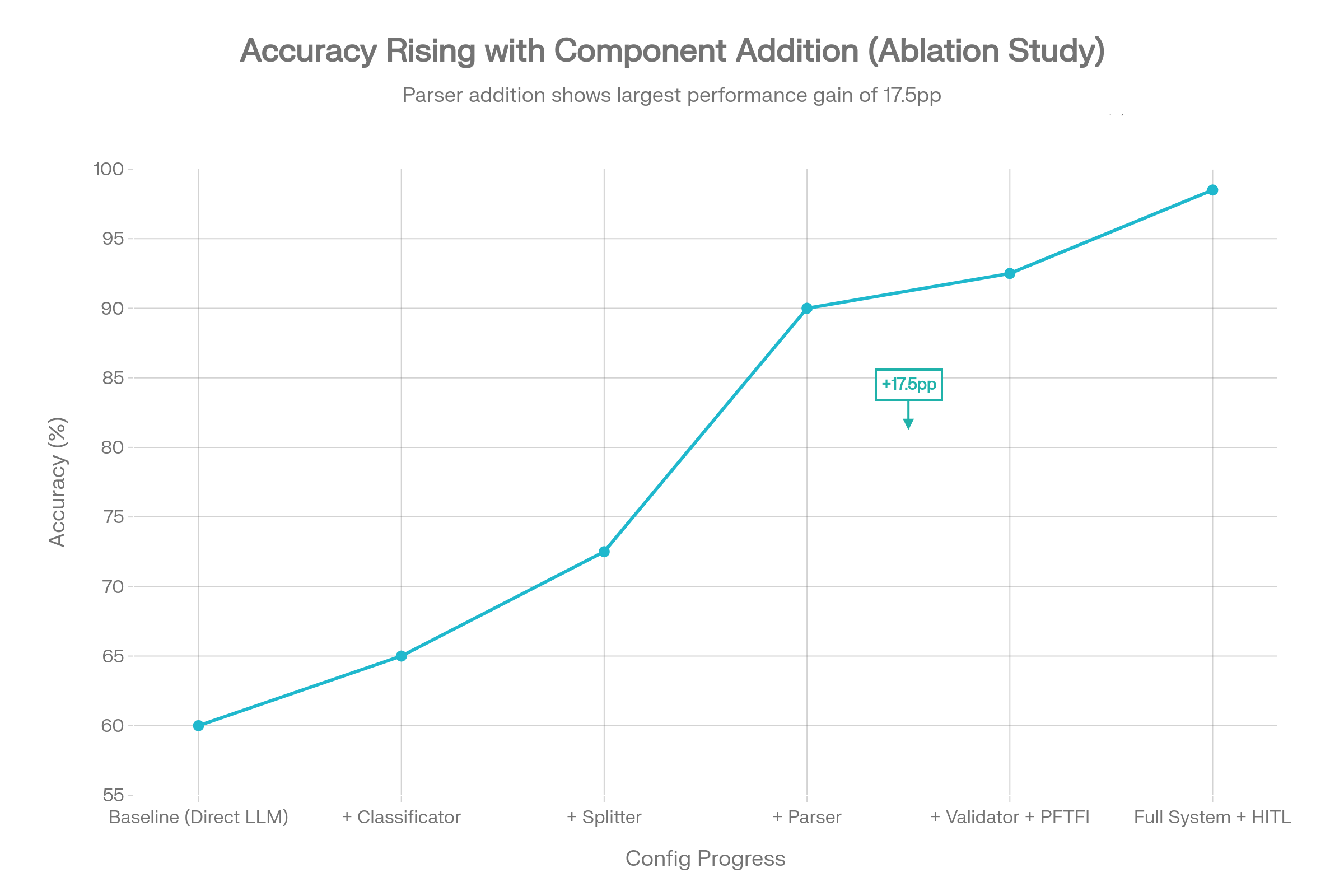}
\caption{Ablation study: incremental accuracy gains per component. Parser Agent contributes the largest improvement (+17.5 pp).}
\label{fig:ablation}
\vspace{-2mm}
\end{figure}

The ablation study was conducted on a stratified subset of 100 documents from the production dataset, comprising 5 documents per each of the 20 supplier/document-type categories present in the deployment. This stratified sampling ensures full coverage of template diversity while maintaining a manageable evaluation set. Table~\ref{tab:ablation} and Figure~\ref{fig:ablation} show the impact of each component on system accuracy. The column ``Categories OK/20'' reports the number of supplier/document-type categories (out of 20) for which all 5 documents were processed correctly on average; non-integer values reflect partial category-level accuracy across the 5-document samples.

\begin{table}[!htbp]
\centering
\caption{Ablation study: Impact of pipeline components on accuracy (100 documents, 5 per each of 20 supplier/document-type categories)}
\label{tab:ablation}
\begin{tabular}{lcc}
\toprule
\textbf{Configuration} & \textbf{Accuracy} & \textbf{Categories OK/20} \\
\midrule
Baseline (Direct LLM on PDF) & 60.0\% & 12/20 \\
+ Classificator Agent & 65.0\% & 13/20 \\
+ Splitter Agent & 72.5\% & 14.5/20 \\
+ Parser Agent & \textbf{90.0\%} & \textbf{18/20} \\
+ Validator + PFTFI (automated) & 92.5\% & 18.5/20 \\
\textbf{Full MADP + HITL} & \textbf{98.5\%} & \textbf{19.7/20} \\
\bottomrule
\end{tabular}
\end{table}

The Parser Agent provides the largest single improvement (+17.5 percentage points), validating our hypothesis that structured preprocessing significantly enhances LLM extraction accuracy. The row ``+ Validator + PFTFI (automated)'' reflects the automated rule-based validation and consistency checks operating without human corrections; the subsequent ``Full MADP + HITL'' row adds human corrections as the feedback source for PFTFI, enabling the full learning loop and reaching 98.5\% accuracy on the ablation subset. This figure is consistent with the 97.0\% full-pipeline automation rate measured on the complete 955-document production deployment (Section~\ref{sec:production}), which constitutes the primary production-validated result of this work.

\subsection{LLM Backend Comparison}
\label{sec:llm-benchmark}

\begin{figure}[h!]
\centering
\vspace{-2mm}
\includegraphics[width=0.75\textwidth]{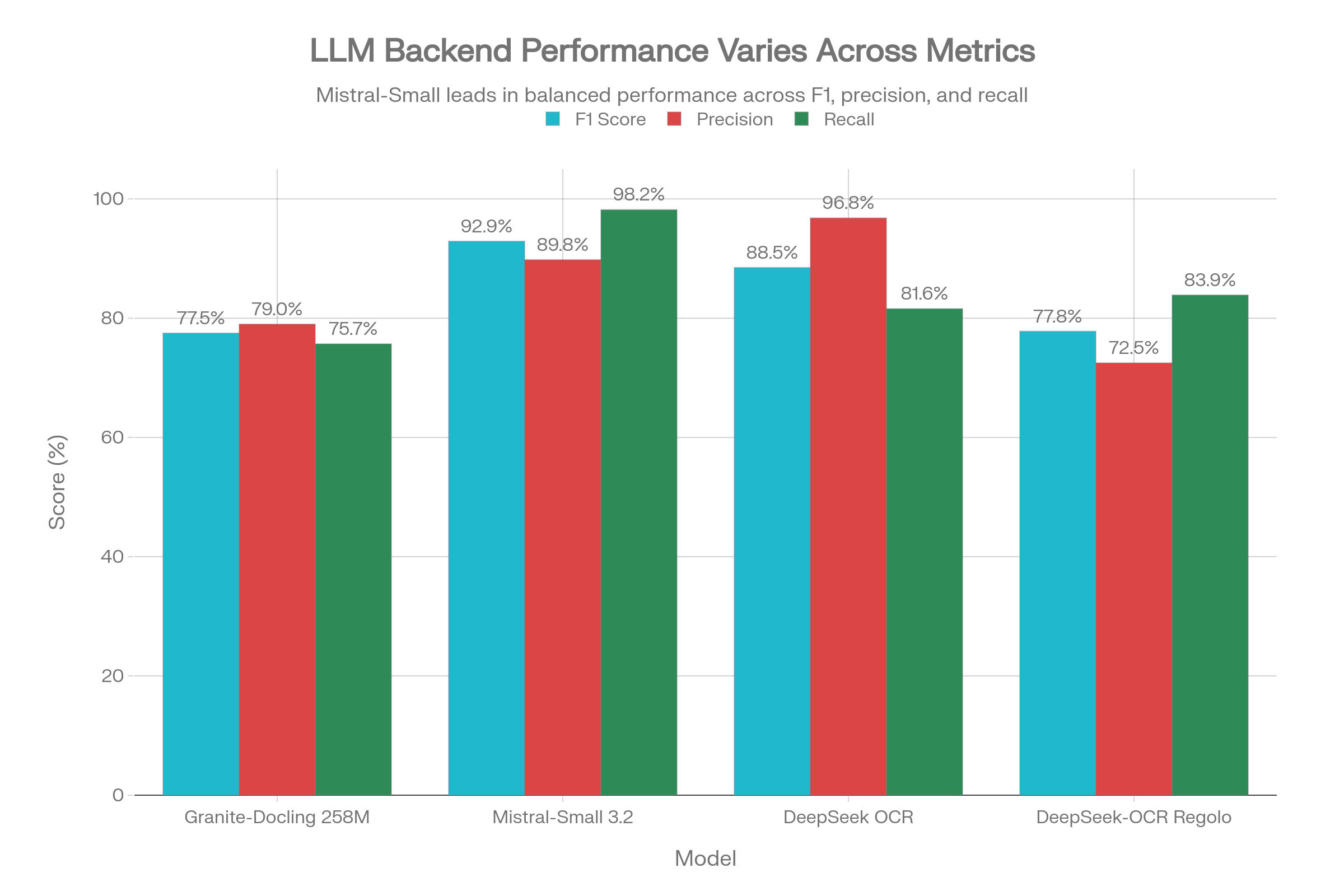}
\caption{LLM backend comparison: Mistral-Small-3.2 leads in F1 (92.9\%) and recall (98.2\%); DeepSeek-OCR leads in precision (96.8\%).}
\label{fig:llmbench}
\vspace{-2mm}
\end{figure}

Table~\ref{tab:llm-benchmark} and Figure~\ref{fig:llmbench} compare four LLM backends on extraction quality and efficiency.

\begin{table}[!htbp]
\centering
\caption{LLM backend comparison for document extraction}
\label{tab:llm-benchmark}
\begin{tabular}{lcccc}
\toprule
\textbf{Model} & \textbf{F1 (\%)} & \textbf{Precision (\%)} & \textbf{Recall (\%)} & \textbf{Time (s/doc)} \\
\midrule
Granite-Docling-258M & 77.5 & 79.0 & 75.7 & 7.76 \\
Mistral-Small-3.2 & \textbf{92.9} & 89.8 & \textbf{98.2} & 17.8 \\
DeepSeek-OCR & 88.5 & \textbf{96.8} & 81.6 & 6.02 \\
DeepSeek-OCR-Regolo & 77.8 & 72.5 & 83.9 & \textbf{3.63} \\
\bottomrule
\end{tabular}
\end{table}

Mistral-Small-3.2 achieves the highest F1 score (92.9\%) with excellent recall (98.2\%), minimizing false negatives at the cost of longer processing time. DeepSeek-OCR offers the best precision (96.8\%), suitable for applications where false positives are costly. DeepSeek-OCR-Regolo provides the fastest inference (3.63s/doc) with acceptable F1 (77.8\%), enabling high-throughput scenarios.

For our production deployment, we use Mistral-Small-3.2 due to its superior accuracy, with DeepSeek-OCR as a fallback for high-volume periods.

\subsection{Operational Efficiency Analysis}

\begin{figure}[h!]
\centering
\vspace{-2mm}
\includegraphics[width=0.75\textwidth]{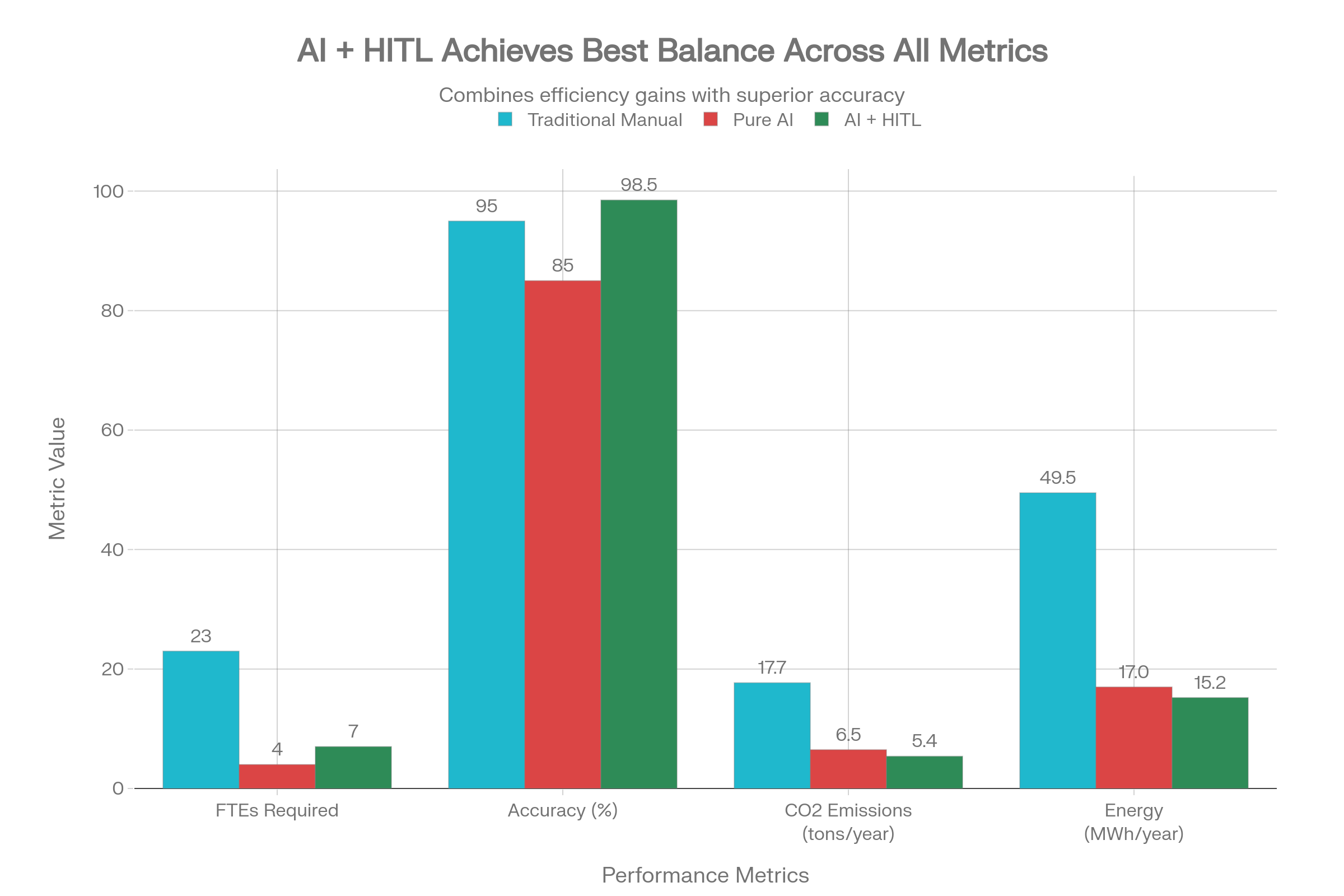}
\caption{Three-scenario comparison: AI+HITL achieves best accuracy (98.5\%), lowest CO\textsubscript{2} (5.4 tons/year), with 7 FTEs.}
\label{fig:operational}
\vspace{-3mm}
\end{figure}

To assess the scalability and economic viability of MADP, we project operational metrics for a representative enterprise use-case scenario of 100{,}000 invoices per year, based on the observed performance characteristics from the 955-document production deployment. Table~\ref{tab:operational} and Figure~\ref{fig:operational} compare MADP against two baseline scenarios: Manual processing, where human operators perform end-to-end document handling including reading, data entry, and validation with minimal automation beyond basic PDF viewing tools; and Pure AI, representing fully automated pipelines with classification, OCR-based parsing, and LLM extraction but without human validation loops. Traditional OCR-based invoice processing systems operating without systematic human review fall into this Pure AI category.

The FTE allocations reflect operational overhead beyond direct document processing activities. Manual processing requires 23 FTE, comprising 20 for data entry and validation and 3 for supervision, training, and quality audits. Pure AI requires 4 FTE for system monitoring and exception handling of documents routed to fallback workflows due to classification failures or system errors (2 FTE), model maintenance and prompt updates (1 FTE), and infrastructure operations and incident response (1 FTE). However, Pure AI achieves only 85\% accuracy, necessitating downstream error correction processes not captured in this FTE count. The AI+HITL configuration requires 7 FTE, allocated to human validation of flagged documents based on the observed 15\% review rate (5 FTE), and system oversight, PFTFI prompt refinement, and continuous improvement activities (2 FTE). This selective human intervention achieves 98.5\% accuracy while avoiding the overhead of reviewing all 100{,}000 invoices manually.

\begin{table}[!htbp]
\centering
\caption{Operational efficiency: 100,000 invoices per year}
\label{tab:operational}
\begin{tabular}{lccc}
\toprule
\textbf{Metric} & \textbf{Manual} & \textbf{Pure AI} & \textbf{AI + HITL} \\
\midrule
Invoices/FTE/year & 4,500 & 25,000 & 15,000 \\
FTEs Required & 23 & 4 & 7 \\
Accuracy & 95\% & 85\% & 98.5\% \\
Human Review Rate & 100\% & 0\% & 15\% \\
Avg. Processing Time & 120s & 6s & 18s \\
\bottomrule
\end{tabular}
\end{table}

The hybrid AI+HITL approach achieves optimal balance: 70\% FTE reduction versus manual processing while maintaining 98.5\% accuracy compared to 85\% for pure AI automation. Strategic human intervention on 15\% of documents provides quality assurance without overwhelming reviewers.

\section{Sustainability Impact Analysis}

We present the first comprehensive sustainability analysis of AI-assisted document processing, quantifying environmental metrics across three scenarios and applying the use-case scenario of 100{,}000 invoices per year to quantify environmental impact at enterprise scale.

\subsection{Methodology}

We calculate CO\textsubscript{2} emissions, energy consumption, and water usage based on established benchmarks for human workers and AI inference operations. For office-based human workers, we adopt a baseline CO\textsubscript{2} footprint of 3.5 kgCO\textsubscript{2}e per worker per day, which accounts for commuting, heating, ventilation, air conditioning, and office equipment~\cite{circularecology2023}. Energy consumption is estimated at approximately 160 kWh per square metre per year for commercial office buildings~\cite{oswbz2017office}, with an allocation of 10 square metres per worker, while water usage includes both direct consumption and facility overhead, amounting to approximately 20 litres per worker per day~\cite{greenly2024}. We assume 220 working days per year for full-time employees.

For AI-based inference, we estimate an energy cost of 0.5 Wh per query for efficient large language models, applying a power usage effectiveness (PUE) factor of 1.5 to account for data centre cooling and infrastructure overhead~\cite{samsi2023carbon}. The corresponding CO\textsubscript{2} intensity is set at 0.3 kgCO\textsubscript{2} per kWh, reflecting the average carbon intensity of the European electricity grid~\cite{patterson2024carbon}. Water consumption associated with evaporative cooling in data centres is estimated at 34 mL per query~\cite{li2024water}. In our document processing pipeline, each invoice requires two LLM queries on average, one for extraction and one for validation.

\subsection{Environmental Impact Results}

Table~\ref{tab:sustainability} presents the full sustainability comparison.

\begin{table}[!htbp]
\centering
\caption{Sustainability impact: 100,000 invoices per year}
\label{tab:sustainability}
\small
\begin{tabular}{lcccc}
\toprule
\textbf{Metric} & \textbf{Manual} & \textbf{Pure AI} & \textbf{AI + HITL} & \textbf{Savings vs Manual} \\
\midrule
\textbf{Annual Totals} & & & & \\
CO2 Emissions (tons) & 17.7 & 3.1 & 5.4 & 12.3 (-69\%) \\
Energy (MWh) & 49.5 & 8.7 & 15.2 & 34.2 (-69\%) \\
Water (m³) & 101.2 & 17.5 & 37.6 & 63.6 (-63\%) \\
\midrule
\textbf{Per Invoice} & & & & \\
CO\textsubscript{2} (gCO\textsubscript{2}e) & 177.1 & 31.0 & 54.4 & 122.8 (-69\%) \\
Energy (Wh) & 494.5 & 87.0 & 152.0 & 342.5 (-69\%) \\
Water (liters) & 1.01 & 0.18 & 0.38 & 0.64 (-63\%) \\
\bottomrule
\end{tabular}
\end{table}

The hybrid AI+HITL approach reduces CO2 emissions by 12.3 tons per year (69\% reduction) relative to the manual baseline. The Pure AI scenario (4 FTE, 85\% accuracy) achieves even lower absolute emissions due to minimal human overhead, but at the cost of substantially reduced accuracy and the need for downstream error correction not captured in these figures. The AI+HITL configuration represents the optimal balance: it achieves the highest accuracy (98.5\%) while reducing environmental impact by 69\% in CO\textsubscript{2} and energy and 63\% in water relative to manual processing. The annual CO\textsubscript{2} reduction of 12.3 tons corresponds to the carbon sequestration of approximately 61 trees planted for one year, or the removal of three passenger cars from circulation for the same period. The energy savings of 34.2 MWh per year are sufficient to power eleven average European homes for one year, while the water savings of 63.6 cubic metres represent 424 days of drinking water for one person.

\subsection{Discussion}

The sustainability analysis indicates that a carefully designed AI pipeline can reduce environmental impact despite the additional computational load introduced by LLM-based processing. The dominant contribution to the observed reductions in CO\textsubscript{2} emissions, energy use, and water consumption arises from the decrease in Full-Time Equivalent (FTE) staffing from 23 to 7 workers, which directly lowers facility-related energy demand, commuting emissions, and associated resource consumption. At the same time, the use of efficient inference configurations for document processing avoids offsetting these gains through excessive data center usage.

The results further suggest that the balance between automation and human oversight is a key determinant of both operational and environmental performance. A fully manual process guarantees high accuracy but is resource-intensive, while a purely automated process reduces FTEs but exhibits lower reliability. The hybrid MADP configuration, which concentrates human intervention on a subset of cases, achieves high accuracy and substantially lower impact per processed invoice. As model architectures and deployment practices continue to improve in terms of energy efficiency~\cite{patterson2024carbon}, it is reasonable to expect additional sustainability gains from future iterations of the framework. In parallel, techniques such as semantic caching and prompt-injection–resilient agentic architectures can further reduce redundant computation while maintaining robustness and security~\cite{gosmar2026promptinjectionmitigationagentic}.

\section{Limitations}

Some limitations of the current work warrant discussion. First, MADP is optimised for invoice processing; adaptation to other document types will require retraining the Classificator CNN and updating extraction prompts for different field structures. Second, the pipeline depends on GPU infrastructure for both the CNN classification and LLM extraction stages, which may present cost barriers for small-scale deployments or organisations with limited computational resources. Third, while the system supports Italian, English, German, and French, expanding to non-Latin scripts will require additional OCR training and LLM evaluation to ensure comparable performance. Finally, new document types or suppliers exhibit higher error rates during a cold-start period until PFTFI accumulates sufficient feedback; we recommend starting with high HITL oversight and gradually reducing manual intervention as accuracy improves.

\section{Future Work}

The current architecture suggests several directions for further work. One line of investigation is the integration of active learning strategies into the PFTFI mechanism, so that human review is focused on documents that are expected to provide the greatest feedback gain. This would refine prompts more efficiently and could further reduce the proportion of invoices requiring manual intervention.

Improving robustness on degraded and atypical documents is another priority. Incorporating additional visual cues—such as logos, signatures, and stamps together with more systematic image preprocessing may lower error rates on low-quality scans and formats that differ from those seen during development.

Privacy-preserving deployment models also deserve attention. Federated approaches, in which organisations share prompt and rule updates rather than raw documents, could allow broader learning from heterogeneous data while respecting confidentiality constraints. Extending the pipeline beyond invoices to contracts, purchase orders, and other business documents will require domain-specific classification and extraction logic, but the modular multi-agent pattern appears transferable.

Finally, the interaction between security, efficiency, and sustainability remains an open question. Combining agentic defences against prompt injection and hallucination with semantic caching and workload-aware routing~\cite{gosmar2025hallucination,gosmar2025promptinjection,gosmar2025sentinel,gosmar2026promptinjectionmitigationagentic,gosmar2025fllm} may reduce redundant or unsafe LLM calls while preserving auditability in regulated settings, and offers a promising direction for future research.

\section{Conclusion}

This paper presented MADP, a specialized-module pipeline with agentic feedback loop for sustainable document processing that combines convolutional neural networks, large language models, and Human-in-the-Loop validation. The empirical evaluation yields two complementary results: a 97.0\% full-pipeline automation rate on the complete 955-document production deployment, and 98.5\% document-level accuracy on a stratified 100-document ablation subset covering all 20 supplier/document-type categories. Operational and sustainability analyses based on a use-case scenario of 100{,}000 invoices per year demonstrate that the architecture achieves approximately 70\% FTE reduction while maintaining high accuracy. In particular, the parser component alone contributes an improvement of 17.5 percentage points in document-level accuracy, confirming the importance of layout-aware representations for downstream LLM-based extraction.

Production deployment statistics covering 955 documents processed through January 2026 show that approximately 97\% were handled entirely by the MADP pipeline, with only 3\% requiring non-AI fallback workflows, further validating the robustness and practical viability of the architecture in operational settings.

The proposed Prompt Fine Tuning with Feedback Inheritance (PFTFI) mechanism demonstrates that it is possible to incorporate human corrections into the system without retraining the underlying models. By capturing error patterns and updating prompts and extraction rules at the agent level, the framework supports continuous improvement during deployment and enables faster adaptation to new suppliers and document variants. The sustainability analysis indicates that the hybrid AI+HITL configuration not only improves operational performance but also reduces annual CO\textsubscript{2} emissions, energy consumption, and water usage compared to a purely manual workflow, with most of the gains attributable to reduced office-related overhead.

\section*{Acknowledgments}
We thank the Tesisquare engineering team for their contributions to the MADP implementation and deployment and the Polytechnic University of Turin, Management Engineering Department for their professional revision.

\bibliographystyle{plain}
\bibliography{references}

\end{document}